\documentclass[a4paper,floatfix,amsmath,amssymb,pre,showpacs,twocolumn,superscriptaddress,nofootinbib]{revtex4-1}
\usepackage{graphicx}
\usepackage{amsmath}
\usepackage{epstopdf}
\usepackage{csquotes}
\usepackage{subcaption}
\date{}
\begin{document}
\title{Investigation of A Collective Decision Making System of Different Neighbourhood-Size Based on Hyper-Geometric Distribution}
\author{Debdipta Goswami}
\affiliation{Department of Electronics and Telecommunication Engineering, Jadavpur University}
\author{Heiko Hamann}
\affiliation{Department of Computer Science, University of Paderborn}
\date{7th August, 2014}
\begin{abstract}
The study of collective decision making system has become the central part of the Swarm-Intelligence Related research in recent years. The most challenging task of modelling a collective decision making system is to develop the macroscopic stochastic equation from its microscopic model. In this report we have investigated the behaviour of a collective decision making system with specified microscopic rules that resemble the chemical reaction and used different group size. Then we ventured to derive a generalized analytical model of a collective-decision system using hyper-geometric distribution.\\
Index Terms-\emph{swarm}; \emph{collective decision making}; \emph{noise}; \emph{group size}; \emph{hyper-geometric distribution}
\end{abstract}
\maketitle


\section{Introduction}
Distributed and decentralized systems that rely on self-organizations to achieve a specific goal usually characterized by non-linear dynamics. They usually involve amplification or positive feedback as well as damping or negative feedback along with several cooperative-competitive interactions to evolve and navigate towards its desired state \cite{ref1}. Due to the non-linearity involved and a great number of microscopic details, these systems are usually difficult to analyze and establishing direct analytical connection between the microscopic and macroscopic model seems infeasible for most of the practical systems \cite{ref2}. The principle challenge in this field to establish a macro-micro link \cite{ref3},\cite{ref4} which has applications in Engineering \cite{ref5}, Sociology, Biology and economics. There are some literature \cite{ref5},\cite{ref6}, \cite{ref7} \& \cite{ref8} proposing an approximation to a macro-micro model that can predice individual swarm trajectories as well as swarm densities. But if the swarm-system is characterized by inhomogeneous spatial distribution, the analysis is even more difficult \cite{ref5} \& \cite{ref8}. 

In this project we focus on self-organizing Collective Decision Making systems (CDM) that usually allow simpler modelling and therefore a good subject to approach for establishing macro-micro link and investigating inhomogeneous spatial distribution. In CDM systems we have a multitudes of choices and each agent can have one opinion at a time. Usually they are initialized to an unordered state without any pattern. Gradually due to self-organization they develop into a pattern. Positive-feedback reinforce the majority decision while negative-feedback prevents the swarm to reach an extreme point. The microscopic rule-sets determine how an agent will modify its opinion w.r.t. the opinions of the neighbouring agents.  CDM are common in natural as well as in artificial system. The natural examples are ant-colonies \cite{ref9} \& \cite{ref10} and marching band formation in locusts \cite{ref11}. In swarm robotics \cite{ref12} \& \cite{ref13} the robustness of CDM systems are often used to generate an efficient model.

Here we have taken a cue from the work of Biancalani \emph{et al} \cite{ref14} and expanded the model following the method described in Hamann \emph{et al} \cite{ref15}. The former one describes a completely noise driven binary CDM with random transition from one decision to the other. There the swarm tends to converge to an undecided state though that can be disturbed after some time step by noise. Hamann \emph{et al} \cite{ref15} used a larger neighbourhood size and majority rule to determine the transition in an agent's decision. In this project, however, even larger neighbourhood size of 5 and 7 has been examined, and all possible combination of majority and minority rules are applied to find out the influence of individual rule in the system trajectory. At last an analytic approach is used to justify the probabilities and link them to the macroscopic behaviour of the trajectory. It's seen that hyper-geometric distribution is applicable to the CDM system like the one investigated here. We focus mainly on the drift term to that allows to predict long term system behaviour.

\section{Model}
Let us consider a swarm of N agents undergoing a CDM process. Each agent is characterized by its current opinion. For simplicity we restrict ourselves to binary decision procedure with two decisions $X_1$ and $X_2$. During CDM process, an agent modifies its opinion w.r.t. the opinions of its neighbouring agents within a specified perception range. The rules are usually governed by majority, i.e. the agent changes its decision in favour of the majority decision of its neighbourhood. The number of agents in its perceptible neighbourhood is referred to as group and the group size $G$ can be different. To eliminate the need of tie-breaker we choose only only odd number as $G$. To preserve the exploration capability of the swarm we can include minority rules also. Sometime the agents can change their opinion randomly under the influence of the noise. We represent the above described microscopic model by a set of chemical reactions that models all possible causes affecting the opinion of an agent-the
reaction schema. The definition of the reaction schema depends on the particular scenario of interest. We give general equations to define the reaction schema:
\begin{equation}
\label{rule}
R_{i,j}^{M,m}: mX_i+MX_j \xrightarrow r (m\mp 1)X_i +(M\pm 1)X_j
\end{equation}
\begin{equation}
\label{noise}
R_{i,j}: X_i\xrightarrow \epsilon X_j
\end{equation}
Equation~(\ref{rule}) defines the transition w.r.t. a majority or minority rule where an agent encounters m numbers of minority decision holder agents and M number of majority-holders. On the other hand equation~(\ref{noise}) defines the random transition in opinion due to noise.
\section{Experimental Findings}
First we have done some experiments with different neighbourhood size to get an empirical understanding of the macroscopic depiction of the CDM system. The whole swarm size and neighbourhood size are always kept as odd numbers to eliminate the possibility of tie-breakers. The swarm-size is kept at $101$ and we have simulated the CDM process with 5 and 7 neighbourhood size. The rules or the reaction schemas are set in an exhaustive manner to find out analyze all possible situations. To implement the CDM practically, we made the use of modified Urn model where the majority rule and minority rule implement positive and negative feedback respectively and we have used Gillespie algorithm, also known as Stochastic Simulation Algorithm (SSA), a Markov-Chain based Monte-Carlo method proved to give statistically correct trajectories for a certain reaction schema.

The first experiment has been done with a group size of 5 and and using two majority and two minority rules given by:
\begin{equation}
\begin{split}
X_1+4X_2\rightarrow5X_2\\
2X_1+3X_2\rightarrow3X_1+2X_2\\
3X_1+2X_2\rightarrow2X_1+3X_2\\
4X_1+X_2\rightarrow5X_1\\
\end{split}
\end{equation}

\begin{figure*}[h]
\centering
  \begin{subfigure}[b]{0.48\textwidth}
 \includegraphics[trim=0cm 0cm 0cm 0cm, clip=true, height=160 pt, width=\textwidth]{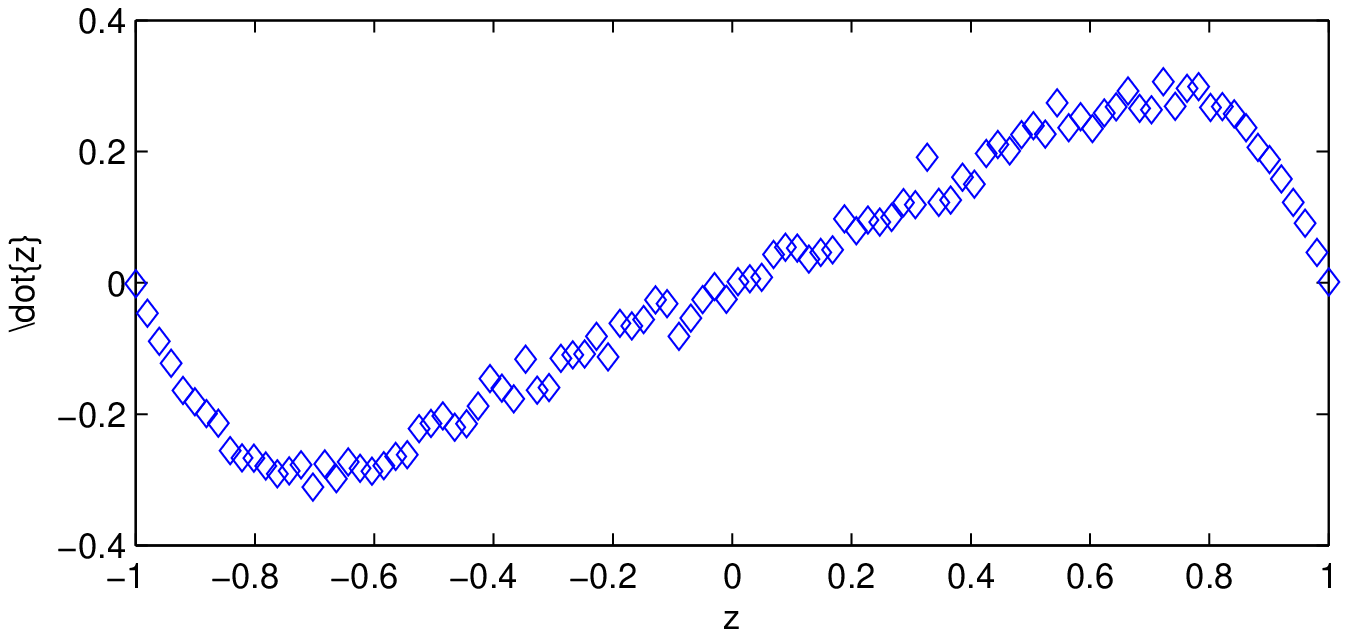}  
 \caption{Plot of $\dot{z}$ vs $z$ without noise for a neighbourhood size of $5$\label{gr_5_noiseN}}
 \end{subfigure}
 \hfill
 \begin{subfigure}[b]{0.48\textwidth}
 \includegraphics[trim=0cm 0cm 0cm 0cm, clip=true, height=160 pt, width=\textwidth]{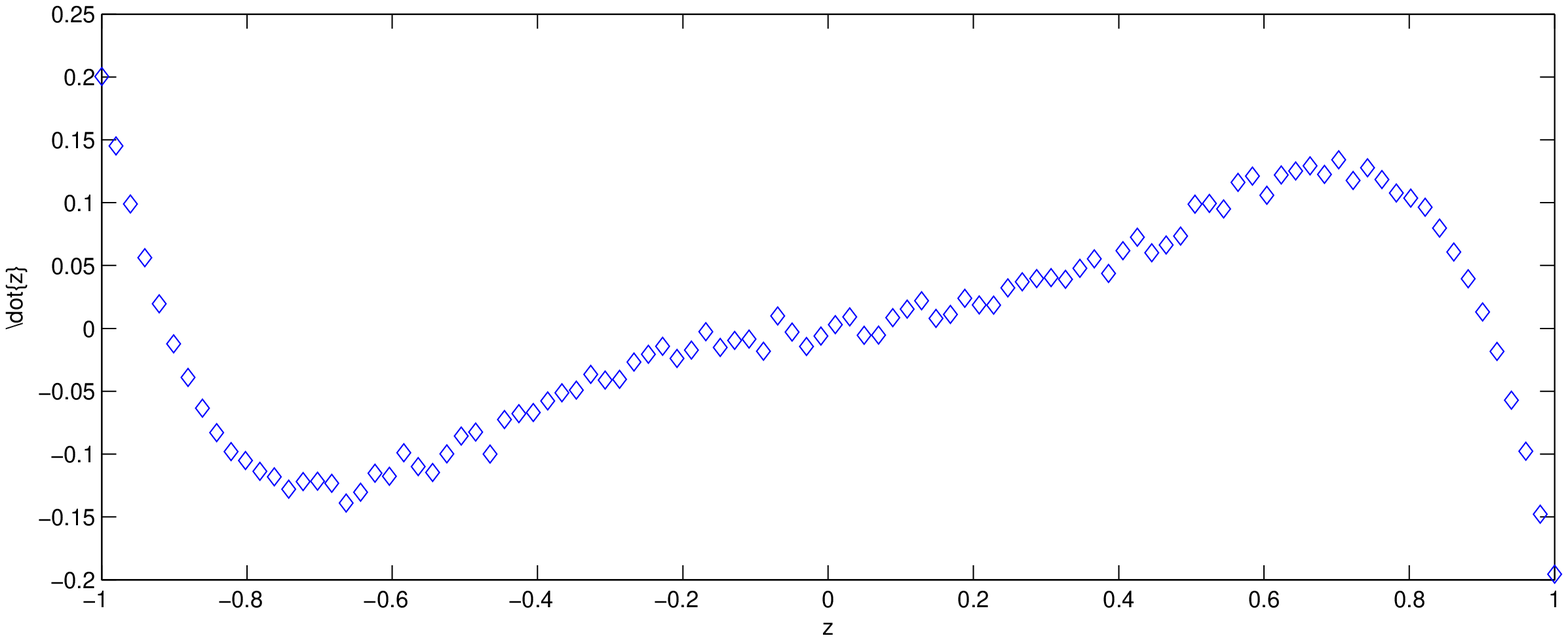}  
 \caption{Plot of $\dot{z}$ vs $z$ with noise for a neighbourhood size of $5$\label{gr_5_noiseY}}
 \end{subfigure}
 \caption{}
\end{figure*}

We have plotted $\dot{z}$ vs $z$ where $z=x_1-x_2$ and $x_1,x_2$ are the fractions of the $X_1$ and $X_2$ entities present in the swarm. The plot thus generated is presented in figure \ref{gr_5_noiseN}. Here there is no noisy transition from $X_1$ to $X_2$. Here we see that there are two stable fixed points at $z=\pm 1$ and one unstable point at $z=0.5$. Now if we introduce noisy transition from $X_1$ to $X_2$ and vice-versa, The positions of the fixed points change and $\pm1$ are not stable any more. The resulting graph is shown in figure \ref{gr_5_noiseY}. 

\begin{figure*}[h]
\centering
 \begin{subfigure}[b]{0.49\textwidth}
 \includegraphics[trim=0cm 0cm 0cm 0cm, clip=true, height=160 pt, width=\textwidth]{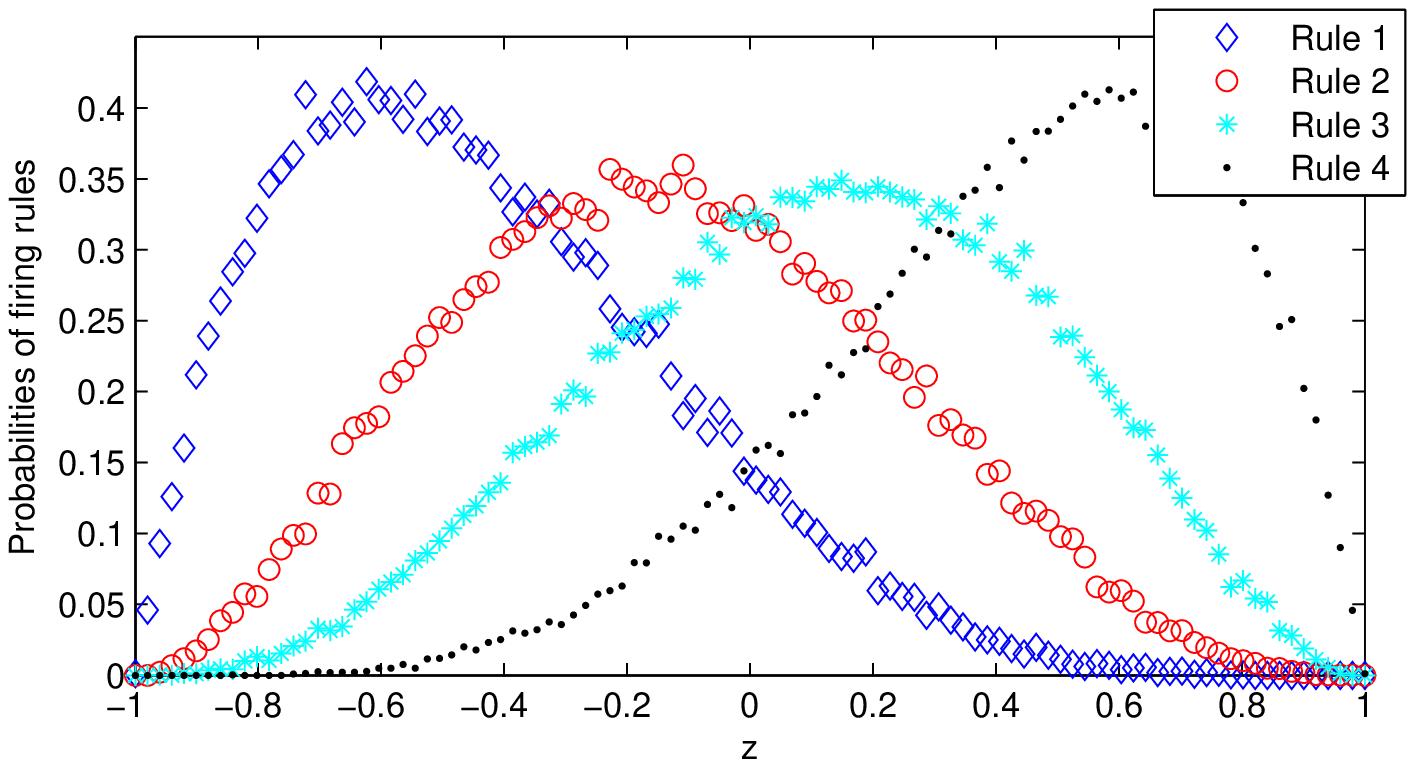}  
 \caption{Plot of the probabilities of firing different rules vs $z$ without noise for a neighbourhood size of $5$\label{gr_5_noiseN_rules}}
 \end{subfigure}
 \hfill
 \begin{subfigure}[b]{0.49\textwidth}
 \includegraphics[trim=0cm 0cm 0cm 0cm, clip=true, height=160 pt, width=\textwidth]{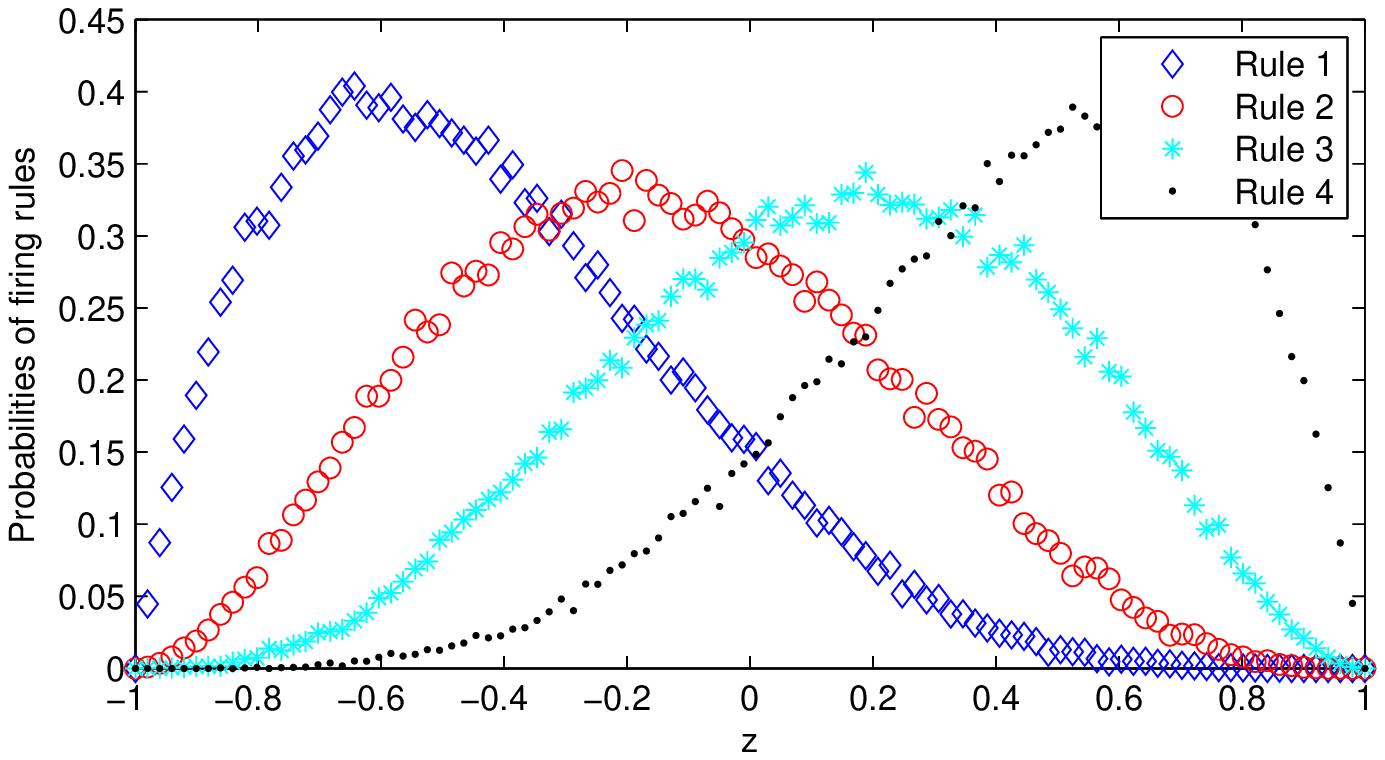}  
 \caption{Plot of the probabilities of firing different rules vs $z$ with noise for a neighbourhood size of $5$\label{gr_5_noiseY_rules}}
 \end{subfigure}
 \caption{}
\end{figure*}

The probabilities of firing the corresponding 4 rules are also plotted w.r.t. $z$ and shown in figures~\ref{gr_5_noiseN_rules} \&~\ref{gr_5_noiseY_rules}. As can be seen from the figures, introduction of noise doesn't change the probabilities of firing of different rules as the noise is merely superimposed with the original system dynamics.

Now we are going to investigate the effect of majority and minority rules for a bigger group size and we have taken group size 7. There are possibility of 6 types of encounters and for them we can set either majority or a minority rule. First we set 4 majority and two minority rules as follows:
\begin{equation}
\label{eq:gr7}
\begin{split}
X_1+6X_2\rightarrow 7X_2\\
2X_1+5X_2\rightarrow X_1+6X_2\\
3X_1+4X_2\rightarrow 4X_1+3X_2\\
4X_1+3X_2\rightarrow 3X_1+4X_2\\
5X_1+2X_2\rightarrow 6X_1+X_2\\
6X_1+X_2\rightarrow 7X_1
\end{split}
\end{equation}

\begin{figure*}[h]
 \begin{subfigure}[b]{0.49\textwidth}
 \includegraphics[trim=0cm 0cm 0cm 0cm, clip=true, height=180 pt, width=\textwidth]{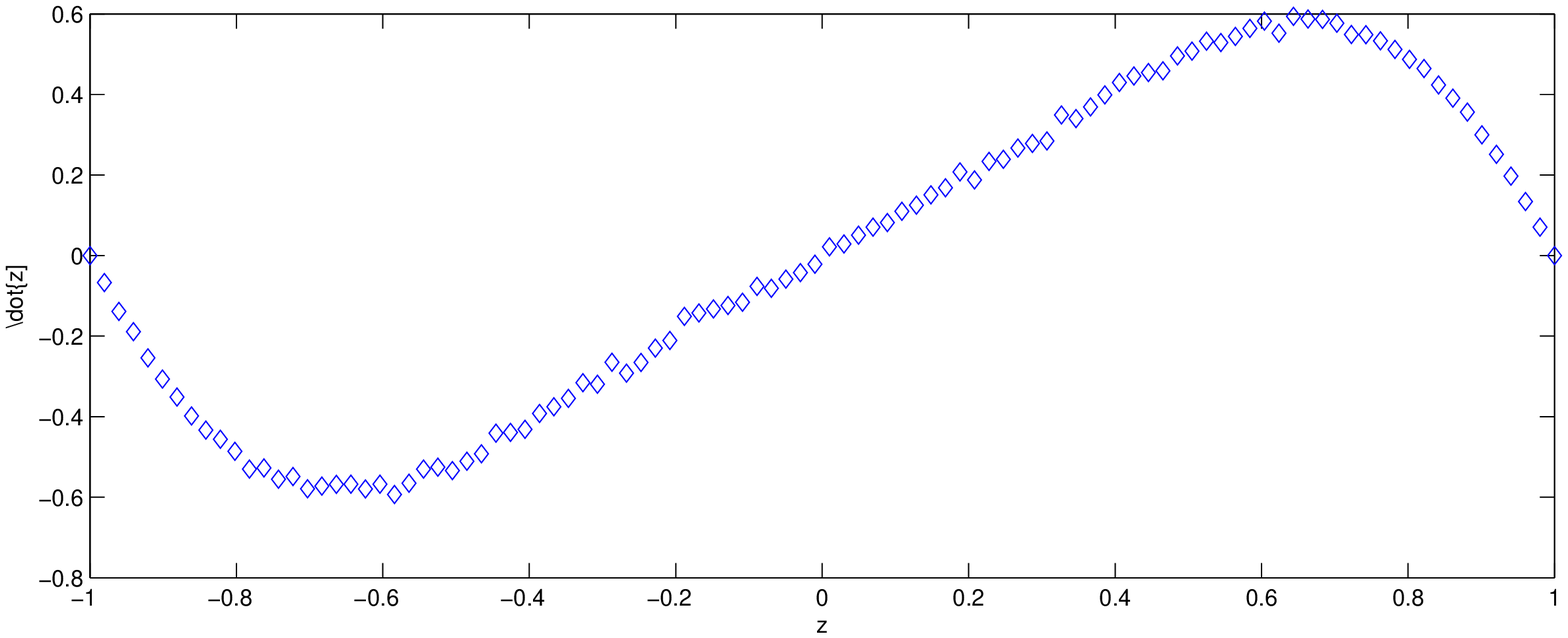}  
 \caption{Plot of $\dot{z}$ vs $z$ without noise for group size 7\label{gr_7_noiseN}}
 \end{subfigure}
 \hfill
 \begin{subfigure}[b]{0.49\textwidth}
 \includegraphics[trim=0cm 0cm 0cm 0cm, clip=true, height=180 pt, width=\textwidth]{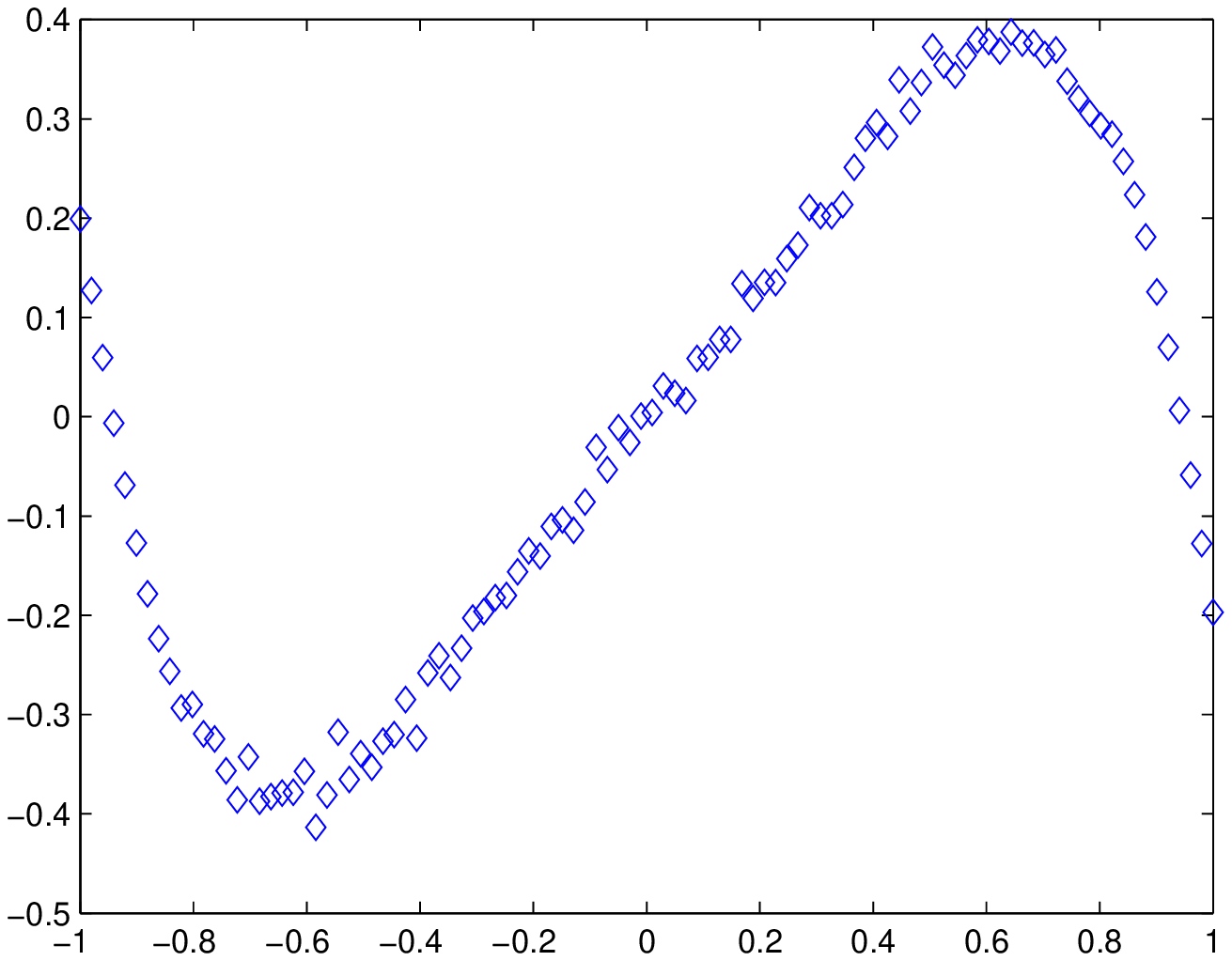}  
 \caption{Plot of $\dot{z}$ vs $z$ with noise for group size 7\label{gr_7_noiseY}}
 \end{subfigure}
 \caption{}
\end{figure*}

The plot of $\dot{z}$ vs $z$ without noise for these set of rules is given in the figure \ref{gr_7_noiseN}. Again if we introduce noise, the critical points are pushed inside from the two extrema as shown in figure \ref{gr_7_noiseY}.

This analysis can be extended further by replacing two of the four majority rules in equation~\ref{eq:gr7} with minority ones. The transition rules will then look like:
\begin{equation}
\begin{split}
X_1+6X_2\rightarrow 7X_2\\
2X_1+5X_2\rightarrow 3X_1+4X_2\\
3X_1+4X_2\rightarrow 4X_1+3X_2\\
4X_1+3X_2\rightarrow 3X_1+4X_2\\
5X_1+2X_2\rightarrow 4X_1+3X_2\\
6X_1+X_2\rightarrow 7X_1
\end{split}
\end{equation}

\begin{figure*}[t]
 \begin{subfigure}[h]{0.49\textwidth}
 \includegraphics[trim=0cm 0cm 0cm 0cm, clip=true, height=180 pt, width=\textwidth]{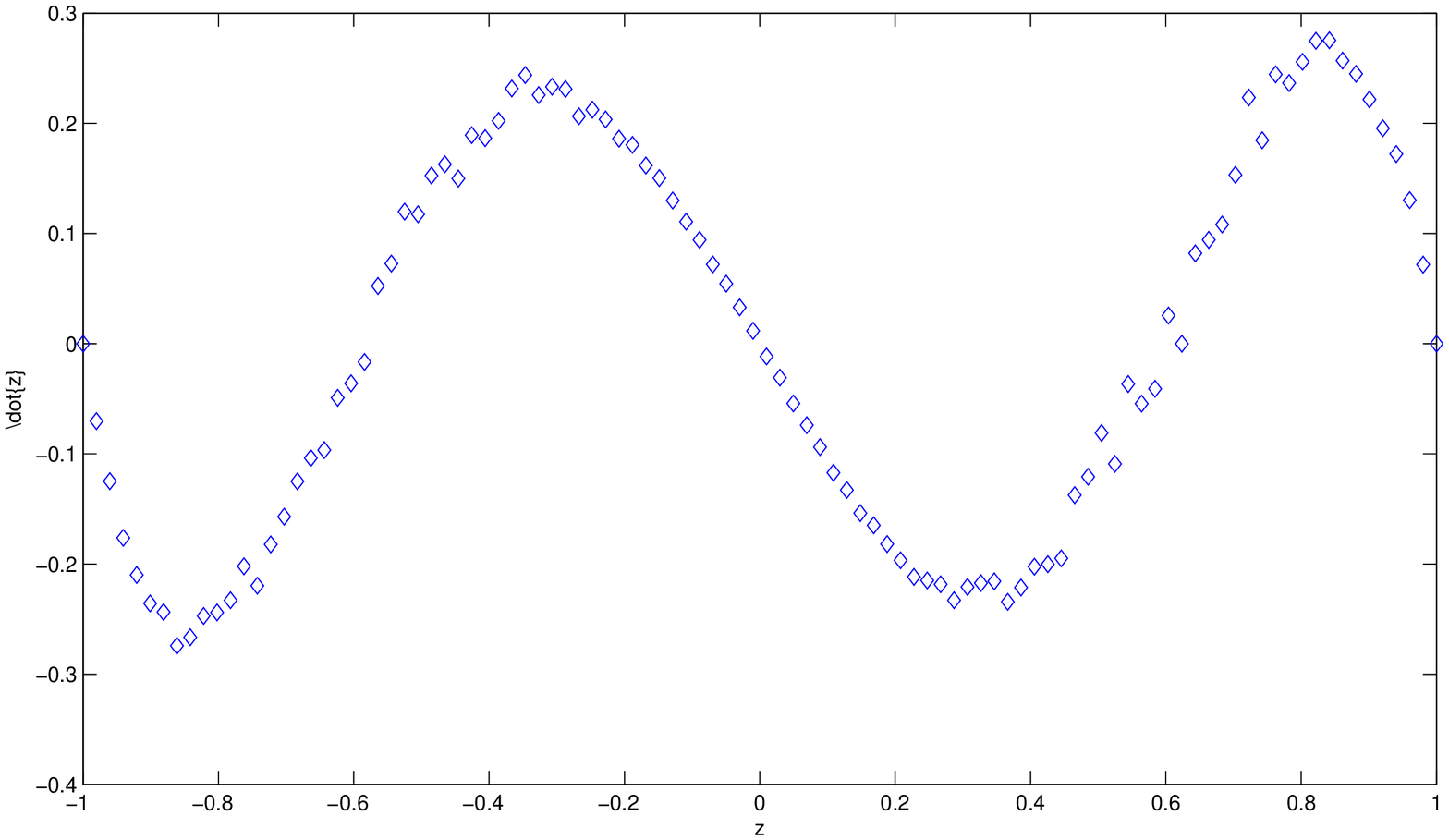}  
 \caption{Plot of $\dot{z}$ vs $z$ without noise for group size 7 with 4 minority rules\label{gr_7_noiseN_minority}}
 \end{subfigure}
 \hfill
 \begin{subfigure}[h]{0.49\textwidth}
 \includegraphics[trim=0cm 0cm 0cm 0cm, clip=true, height=180 pt, width=\textwidth]{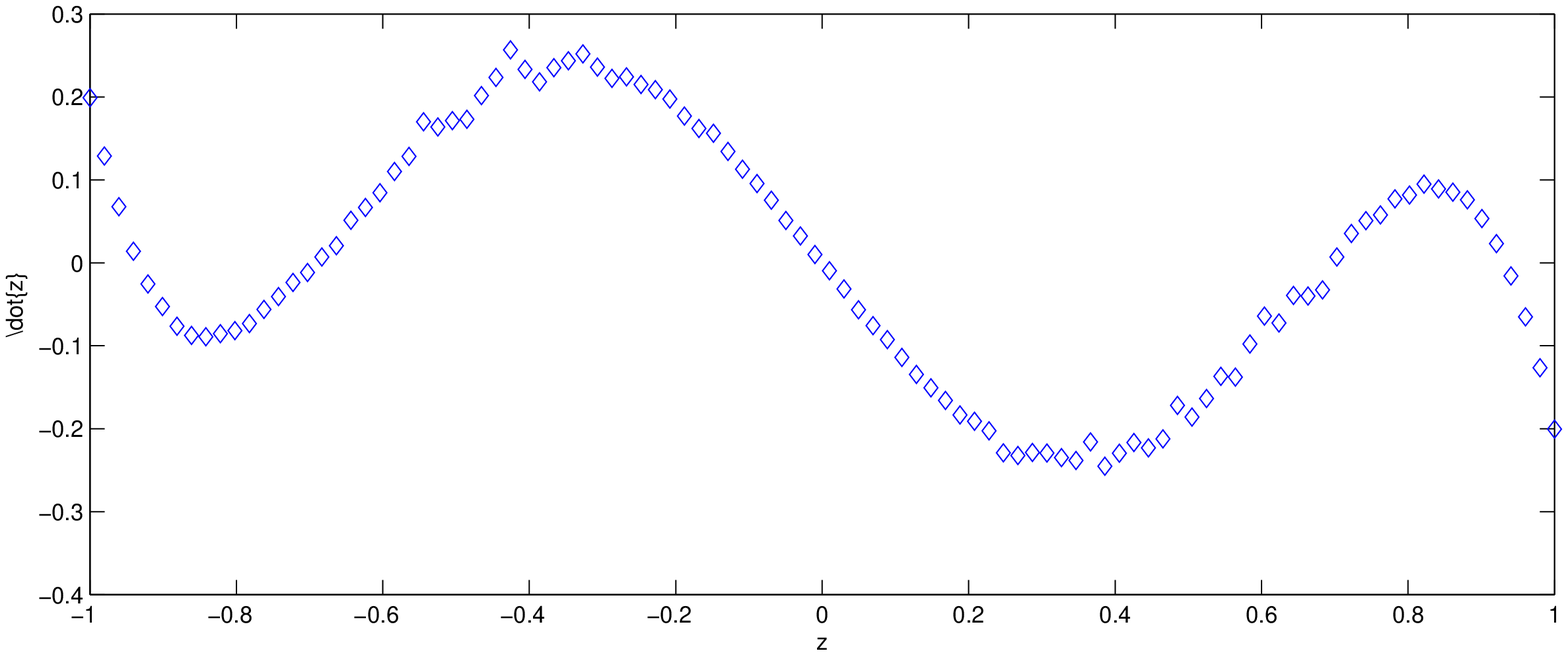}  
 \caption{Plot of $\dot{z}$ vs $z$ with noise for group size 7 with 4 minority rules\label{gr_7_noiseY_minority}}
 \end{subfigure}
 \caption{}
\end{figure*}

For these rules the plot of $\dot{z}$ vs $z$ without noise is shown in figure~\ref{gr_7_noiseN_minority}. There are two new critical points have been introduced. The effect of noise is the same for that of the previous cases. The two extreme points $z=\pm 1$ cease to become the stable critical point with the advent of noise and new stable critical points formed at the pints greater than $z=-1$ and less than $z=1$. The plot with noise is also shown in the figure~\ref{gr_7_noiseY_minority}.

%

The probabilities of firing different rules do not depend on the noise and same for both the cases: with or without noise. It does not change with the change in majority or minority rules too. The corresponding plot of the probabilities w.r.t. $z$ is given in the figure~\ref{gr_7_noiseN_rules}. 

\begin{figure}[h]
 \includegraphics[trim=0cm 0cm 0cm 0cm, clip=true, height=180 pt, width=0.5\textwidth]{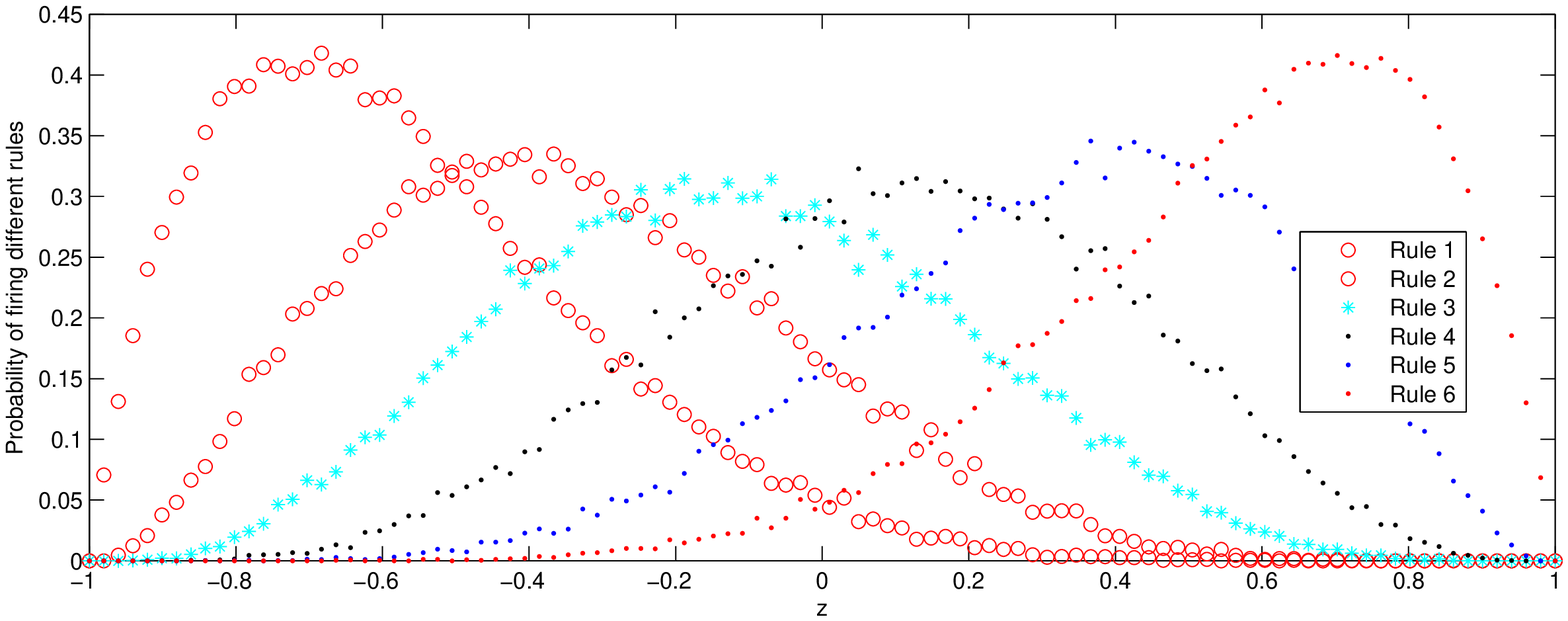}  
 \caption{Plot of the probabilities of firing different rules vs $z$ for a neighbourhood-size of $7$\label{gr_7_noiseN_rules}}
 \label{gr_7_noiseN_rules}
\end{figure}

An interesting incident takes place if we change all the 6 rules into minority ones. The the two extrema $z=\pm 1$ become unstable fixed point rather than stable ones; and the point $z=0$ becomes the stable fixeed point. It can be intuitively explained since with all minority rules, the swarm at the extreme points will always tend to move out from there by the application of a minority rule. The plot of $\dot{z}$ w.r.t. $z$ for all minority rules is shown in the figure~\ref{gr7_all_minority}.
\begin{figure}[t]
 \includegraphics[trim=0cm 0cm 0cm 0cm, clip=true, height=180 pt, width=0.5\textwidth]{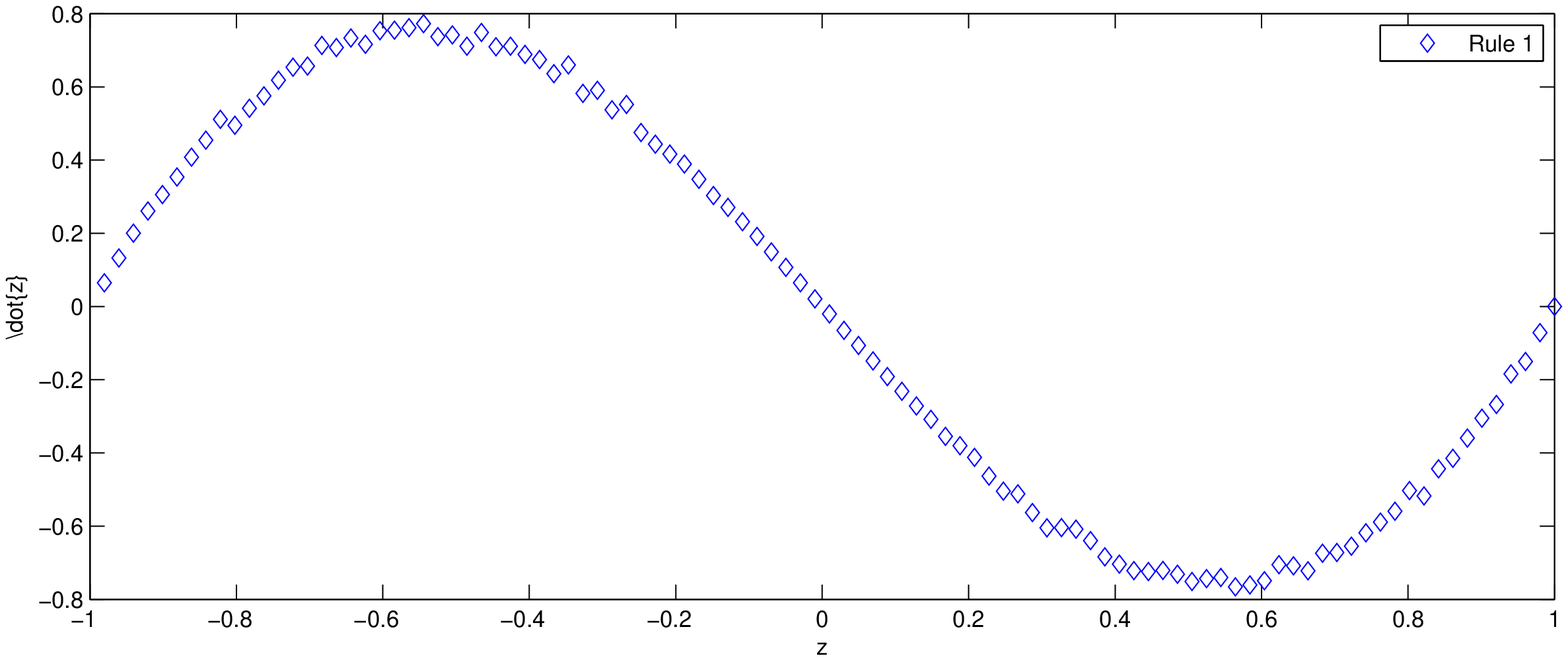}  
 \caption{Plot of $\dot{z}$ vs $z$ without noise for group size 7 with all minority rules\label{gr7_all_minority}}
 \label{gr7_all_minority}
\end{figure}

Now we proceed to experiment further by testing all the possible combination of majority and minority rules. As there should be symmetry, there are 8 different combinations possible. We are writing down the rules for 8 combinations below.
\begin{equation}
\label{MMM}
\begin{split}
X_1+6X_2\rightarrow 7X_2\\
2X_1+5X_2\rightarrow X_1+6X_2\\
3X_1+4X_2\rightarrow 2X_1+5X_2\\
4X_1+3X_2\rightarrow 5X_1+2X_2\\
5X_1+2X_2\rightarrow 6X_1+X_2\\
6X_1+X_2\rightarrow 7X_1
\end{split}
\end{equation}

\begin{equation}
\label{MMm}
\begin{split}
X_1+6X_2\rightarrow 7X_2\\
2X_1+5X_2\rightarrow X_1+6X_2\\
3X_1+4X_2\rightarrow 4X_1+3X_2\\
4X_1+3X_2\rightarrow 3X_1+4X_2\\
5X_1+2X_2\rightarrow 6X_1+X_2\\
6X_1+X_2\rightarrow 7X_1
\end{split}
\end{equation}

\begin{equation}
\label{Mmm}
\begin{split}
X_1+6X_2\rightarrow 7X_2\\
2X_1+5X_2\rightarrow 3X_1+4X_2\\
3X_1+4X_2\rightarrow 4X_1+3X_2\\
4X_1+3X_2\rightarrow 3X_1+4X_2\\
5X_1+2X_2\rightarrow 4X_1+3X_2\\
6X_1+X_2\rightarrow 7X_1
\end{split}
\end{equation}

\begin{equation}
\label{MmM}
\begin{split}
X_1+6X_2\rightarrow 7X_2\\
2X_1+5X_2\rightarrow 3X_1+4X_2\\
3X_1+4X_2\rightarrow 2X_1+5X_2\\
4X_1+3X_2\rightarrow 5X_1+2X_2\\
5X_1+2X_2\rightarrow 4X_1+3X_2\\
6X_1+X_2\rightarrow 7X_1
\end{split}
\end{equation}

\begin{equation}
\label{mMm}
\begin{split}
X_1+6X_2\rightarrow 2X_1+5X_2\\
2X_1+5X_2\rightarrow X_1+6X_2\\
3X_1+4X_2\rightarrow 4X_1+3X_2\\
4X_1+3X_2\rightarrow 3X_1+4X_2\\
5X_1+2X_2\rightarrow 6X_1+X_2\\
6X_1+X_2\rightarrow 5X_1+2X_2
\end{split}
\end{equation}

\begin{equation}
\label{mMM}
\begin{split}
X_1+6X_2\rightarrow 2X_1+5X_2\\
2X_1+5X_2\rightarrow X_1+6X_2\\
3X_1+4X_2\rightarrow 2X_1+5X_2\\
4X_1+3X_2\rightarrow 5X_1+2X_2\\
5X_1+2X_2\rightarrow 6X_1+X_2\\
6X_1+X_2\rightarrow 5X_1+2x_2
\end{split}
\end{equation}

\begin{equation}
\label{mmM}
\begin{split}
X_1+6X_2\rightarrow 2X_1+5X_2\\
2X_1+5X_2\rightarrow 3X_1+4X_2\\
3X_1+4X_2\rightarrow 2X_1+5X_2\\
4X_1+3X_2\rightarrow 5X_1+2X_2\\
5X_1+2X_2\rightarrow 4X_1+3X_2\\
6X_1+X_2\rightarrow 5X_1+2X_2
\end{split}
\end{equation}

\begin{equation}
\label{mmm}
\begin{split}
X_1+6X_2\rightarrow 2X_1+5X_2\\
2X_1+5X_2\rightarrow 3X_1+4X_2\\
3X_1+4X_2\rightarrow 4X_1+3X_2\\
4X_1+3X_2\rightarrow 3X_1+4X_2\\
5X_1+2X_2\rightarrow 4X_1+3X_2\\
6X_1+X_2\rightarrow 5X_1+2X_2
\end{split}
\end{equation}

Here equation~\ref{MMM} and~\ref{mmm} describes all majority and minority rules and there are other 6 rules possible as indicated in the transition equations. Now we have tried to plot $\dot{z}$ vs $z$ for each case without noise to get a clear picture. W.r.t. majority vs minority rule we can see that equations~\ref{MMM} and~\ref{mmm}, \ref{MMm} and~\ref{mmM}, \ref{MmM} and \ref{mMm},~\ref{Mmm} and~\ref{mMM} are mutually opposite to each other. As we see in the plots that the $\dot{z}$ vs $z$ curves are also just opposite in sign for those cases.
\begin{figure*}[t]
\centering
  \begin{subfigure}[b]{0.48\textwidth}
 \includegraphics[trim=0cm 0cm 0cm 0cm, clip=true, height=160 pt, width=\textwidth]{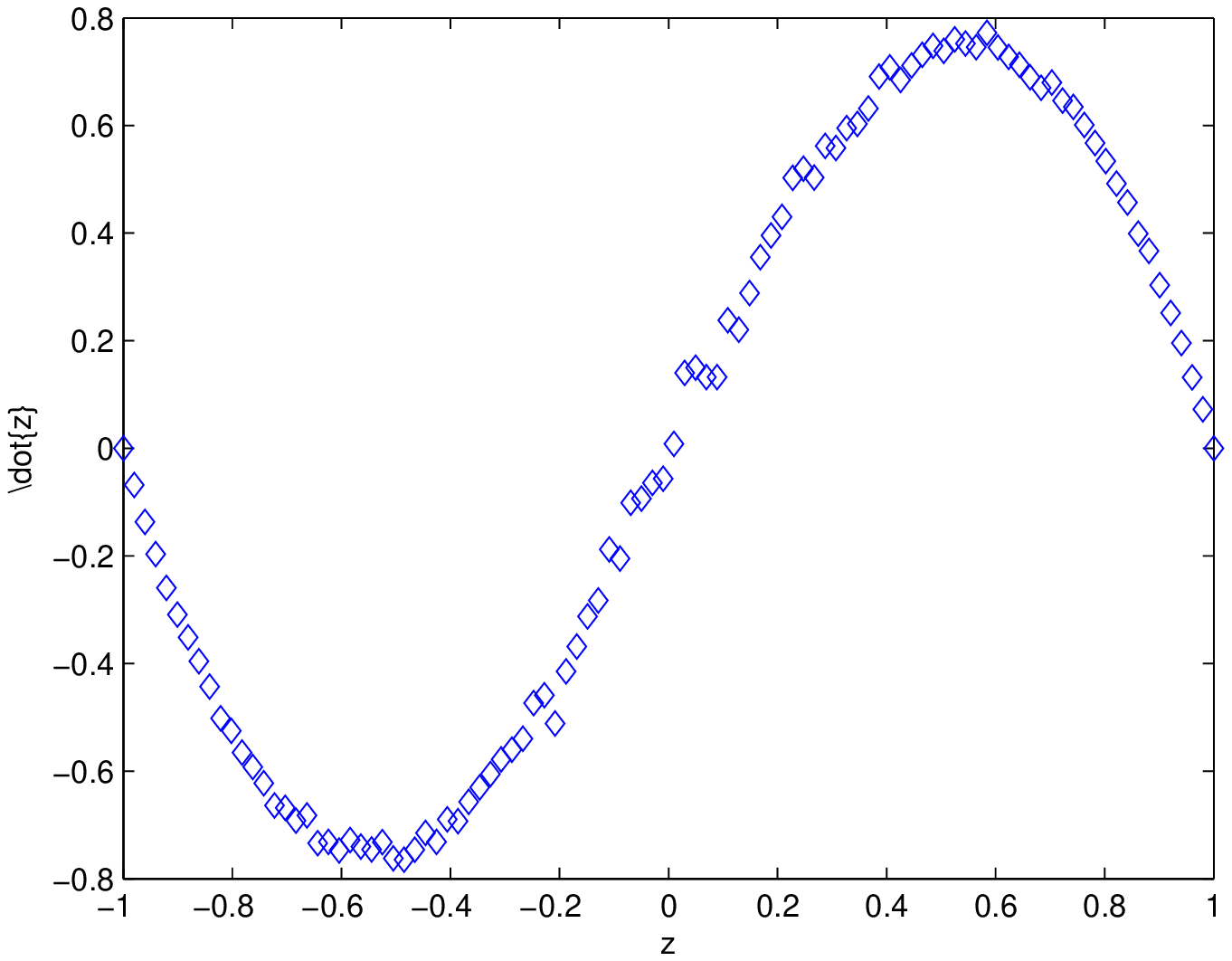}  
 \caption{Plot of $\dot{z}$ vs $z$ without noise according to rule~\ref{MMM}\label{fig:MMM}}
 \end{subfigure}
 \hfill
 \begin{subfigure}[b]{0.48\textwidth}
 \includegraphics[trim=0cm 0cm 0cm 0cm, clip=true, height=160 pt, width=\textwidth]{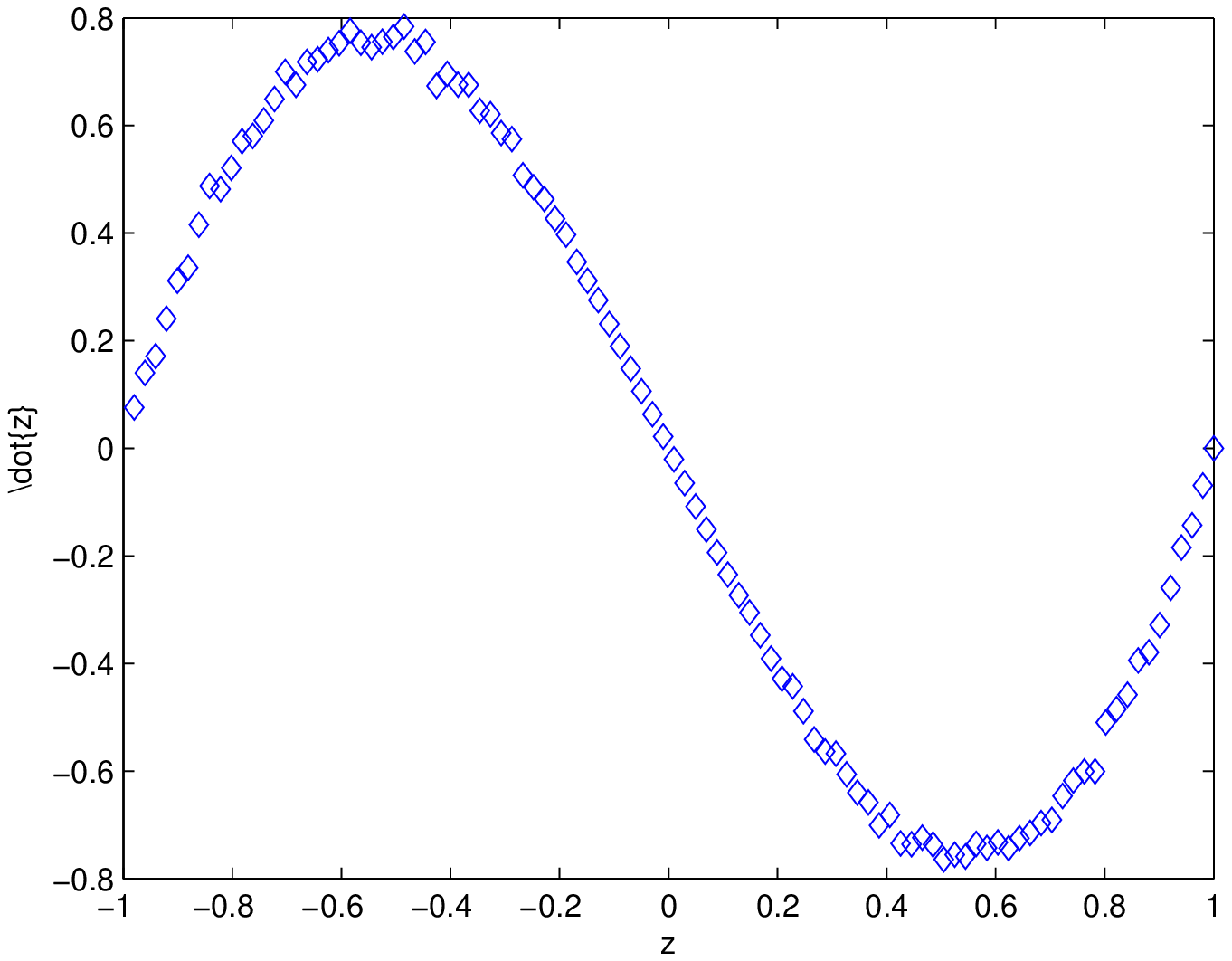}  
 \caption{Plot of $\dot{z}$ vs $z$ without noise according to rule~\ref{mmm}\label{fig:mmm}}
 \end{subfigure}
 \caption{}
\end{figure*}

\begin{figure*}[ht]
\centering
  \begin{subfigure}[h]{0.48\textwidth}
 \includegraphics[trim=0cm 0cm 0cm 0cm, clip=true, height=160 pt, width=\textwidth]{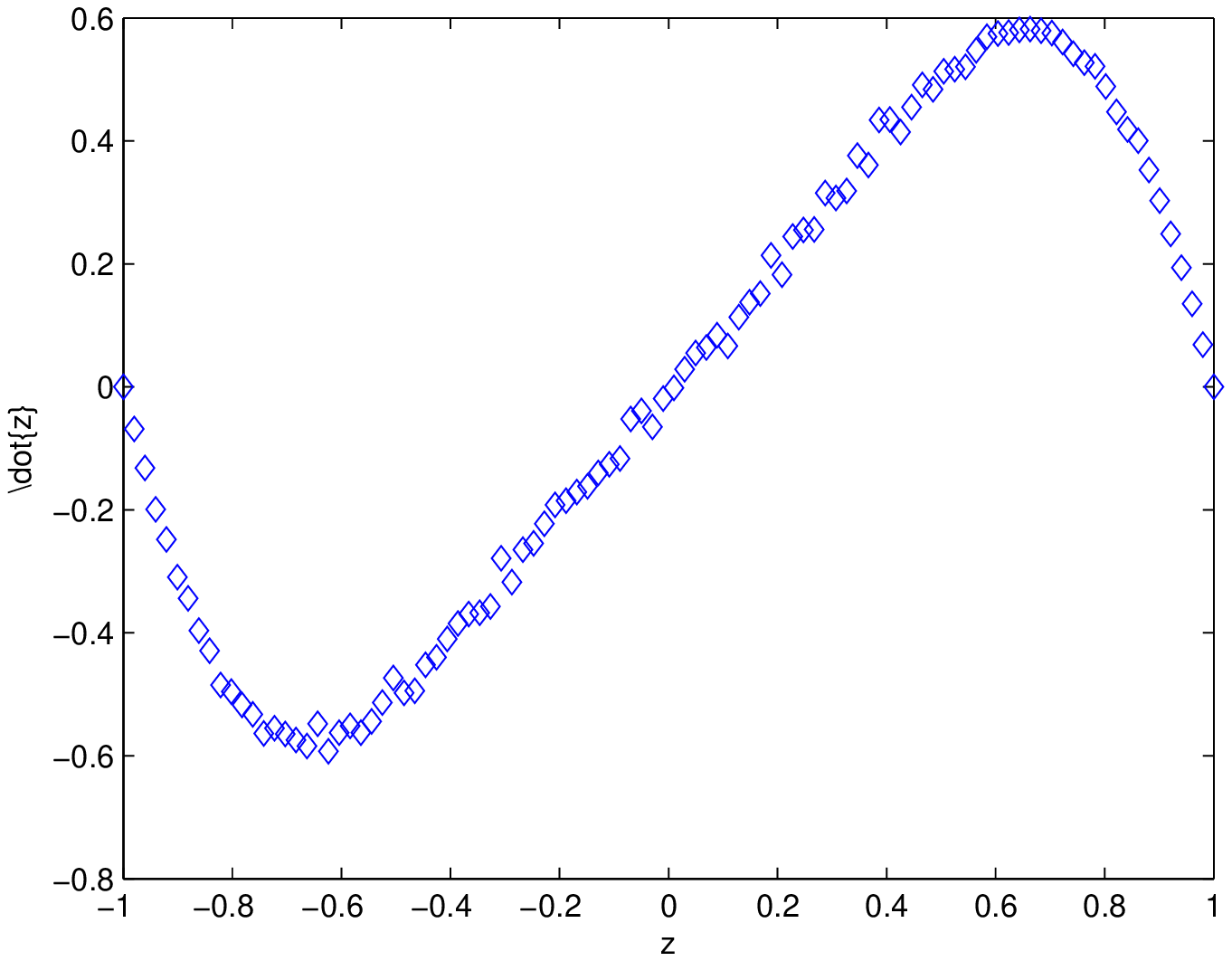}  
 \caption{Plot of $\dot{z}$ vs $z$ without noise according to rule~\ref{MMm}\label{fig:MMm}}
 \end{subfigure}
 \hfill
 \begin{subfigure}[h]{0.48\textwidth}
 \includegraphics[trim=0cm 0cm 0cm 0cm, clip=true, height=160 pt, width=\textwidth]{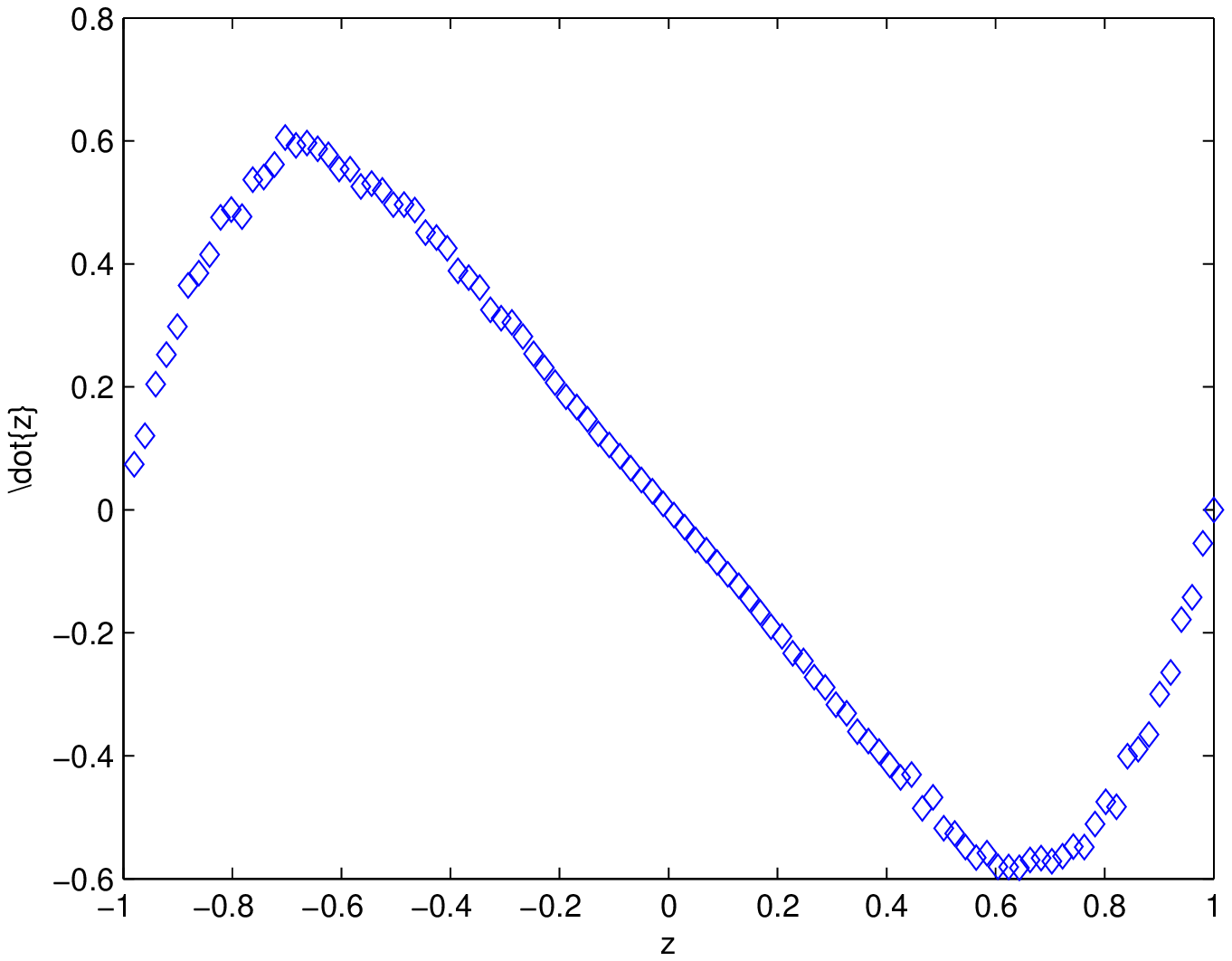}  
 \caption{Plot of $\dot{z}$ vs $z$ without noise according to rule~\ref{mmM}\label{fig:mmM}}
 \end{subfigure}
 \caption{}
\end{figure*}

\begin{figure*}[h]
\centering
  \begin{subfigure}[b]{0.48\textwidth}
 \includegraphics[trim=0cm 0cm 0cm 0cm, clip=true, height=180 pt, width=\textwidth]{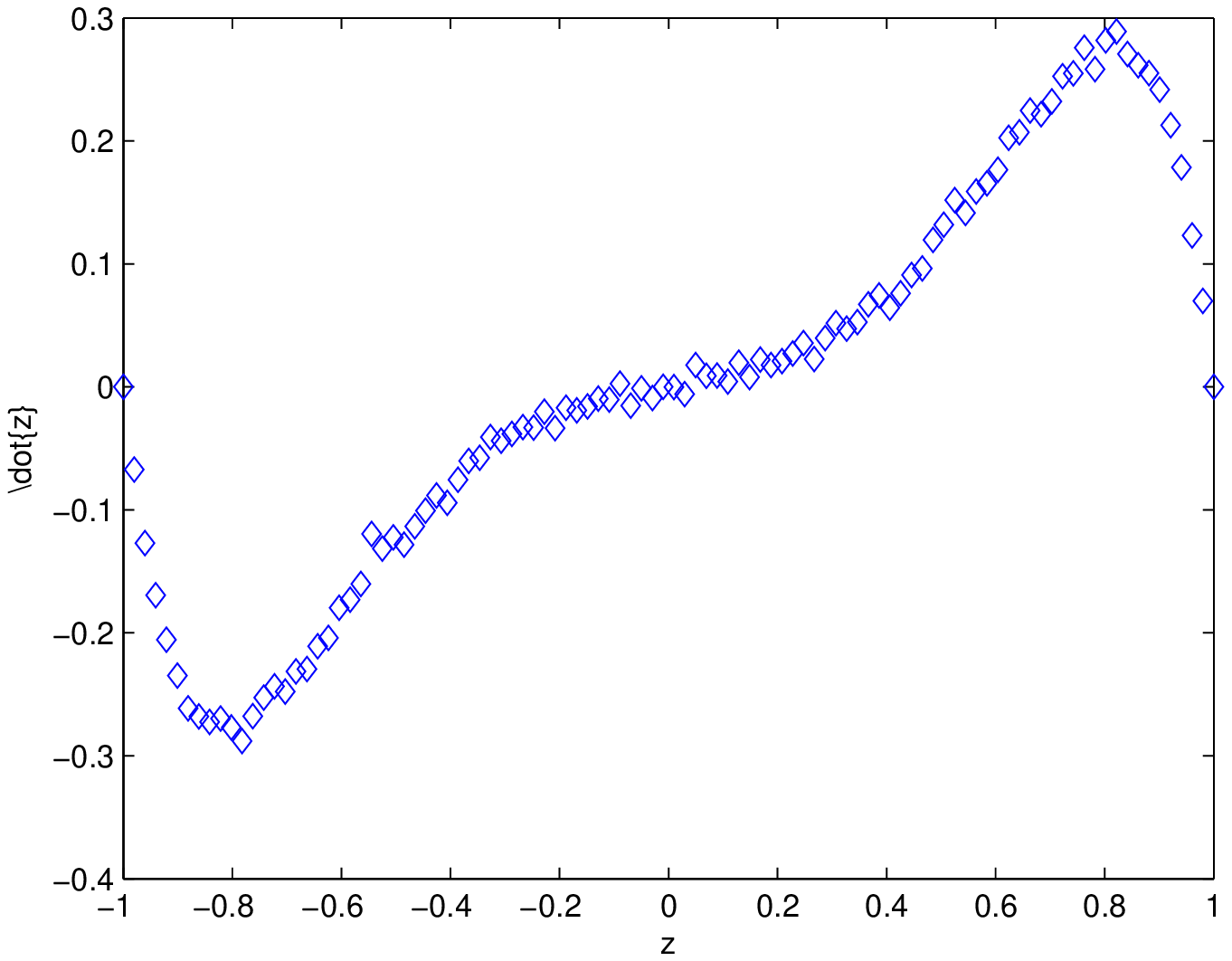}  
 \caption{Plot of $\dot{z}$ vs $z$ without noise according to rule~\ref{MmM}\label{fig:MmM}}
 \end{subfigure}
 \hfill
 \begin{subfigure}[b]{0.48\textwidth}
 \includegraphics[trim=0cm 0cm 0cm 0cm, clip=true, height=180 pt, width=\textwidth]{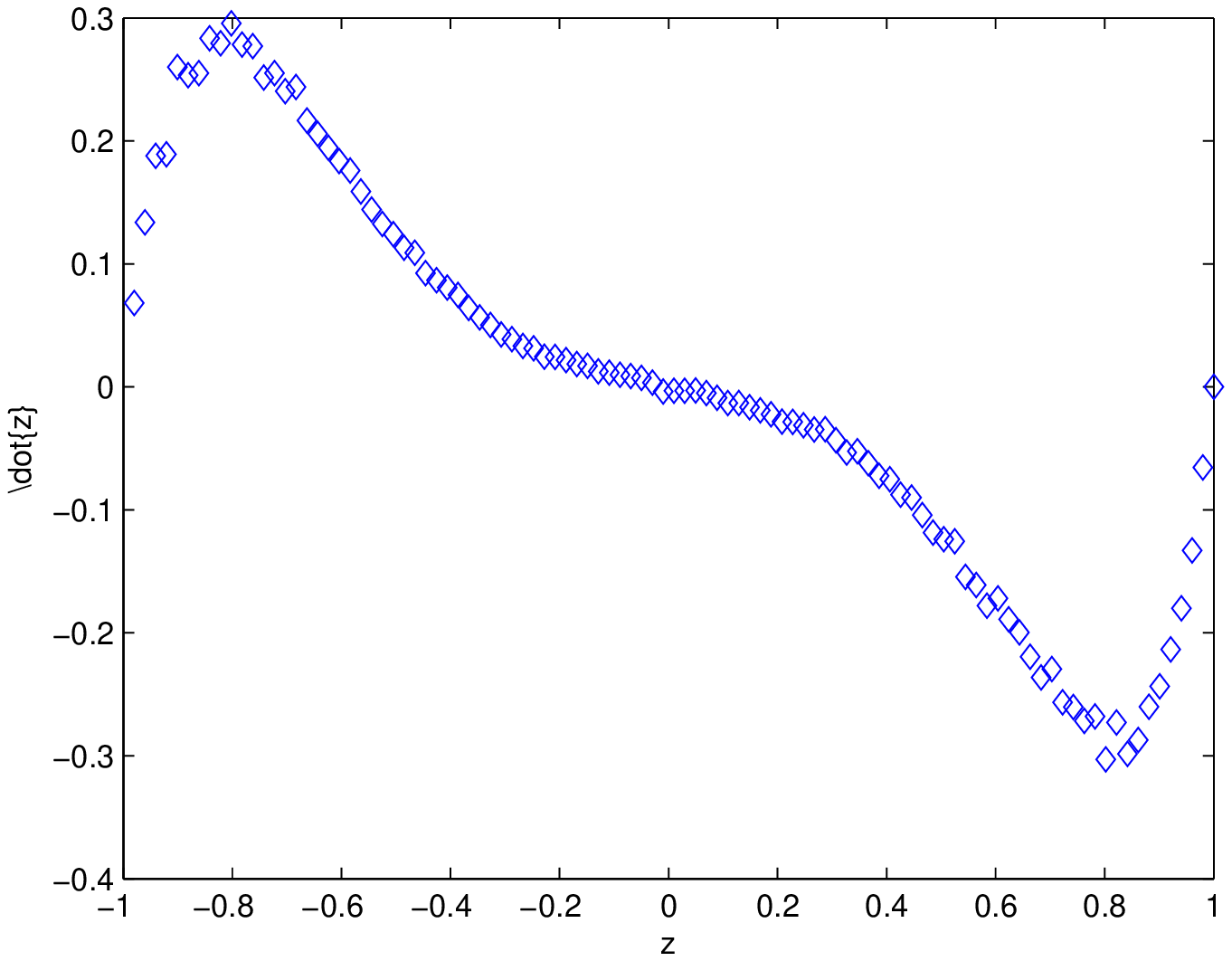}  
 \caption{Plot of $\dot{z}$ vs $z$ without noise according to rule~\ref{mMm}\label{fig:mMm}}
 \end{subfigure}
 \caption{}
\end{figure*}

\begin{figure*}[b]
\centering
  \begin{subfigure}[b]{0.48\textwidth}
 \includegraphics[trim=0cm 0cm 0cm 0cm, clip=true, height=180 pt, width=\textwidth]{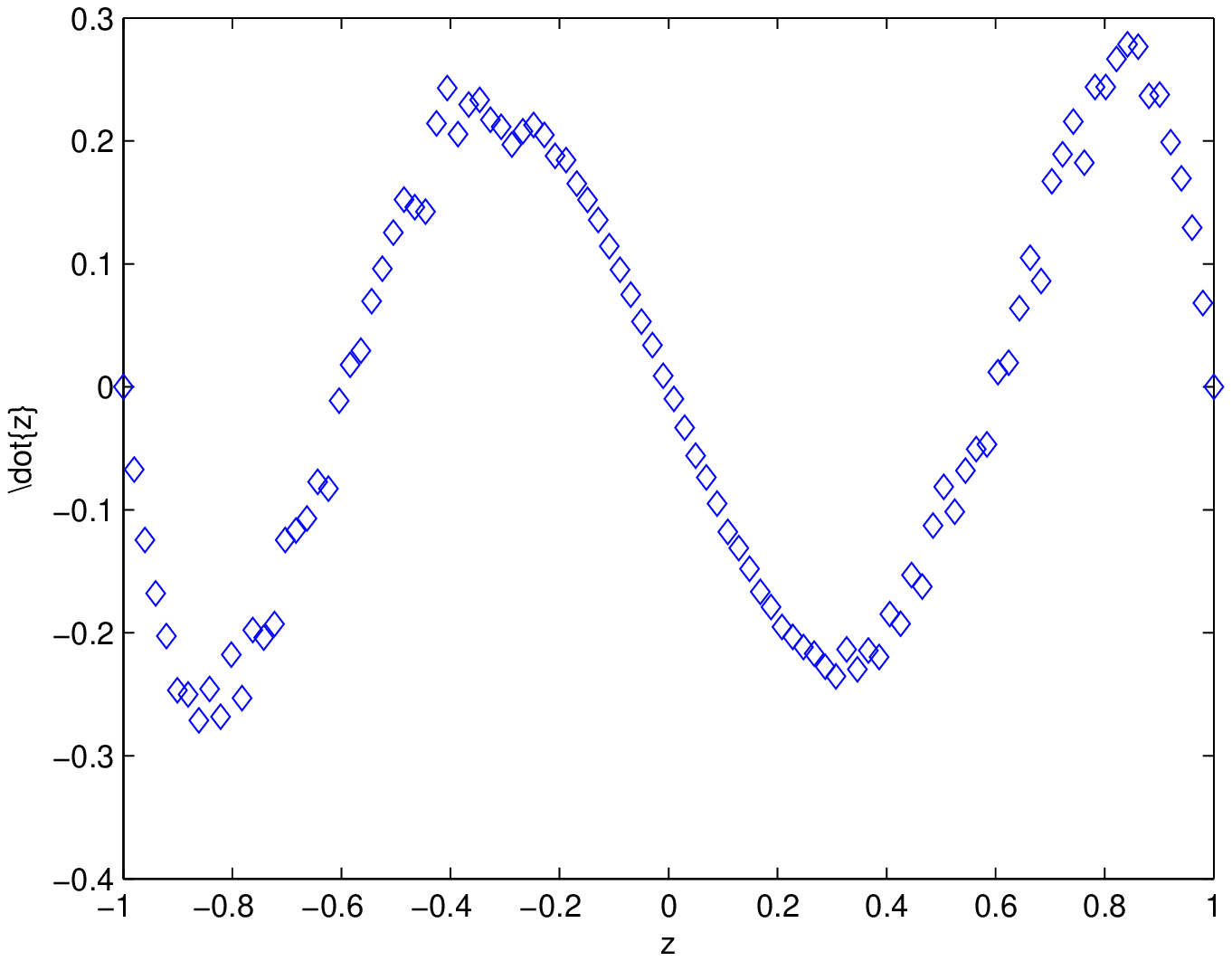}  
 \caption{Plot of $\dot{z}$ vs $z$ without noise according to rule~\ref{Mmm}\label{fig:Mmm}}
 \end{subfigure}
 \hfill
 \begin{subfigure}[b]{0.48\textwidth}
 \includegraphics[trim=0cm 0cm 0cm 0cm, clip=true, height=180 pt, width=\textwidth]{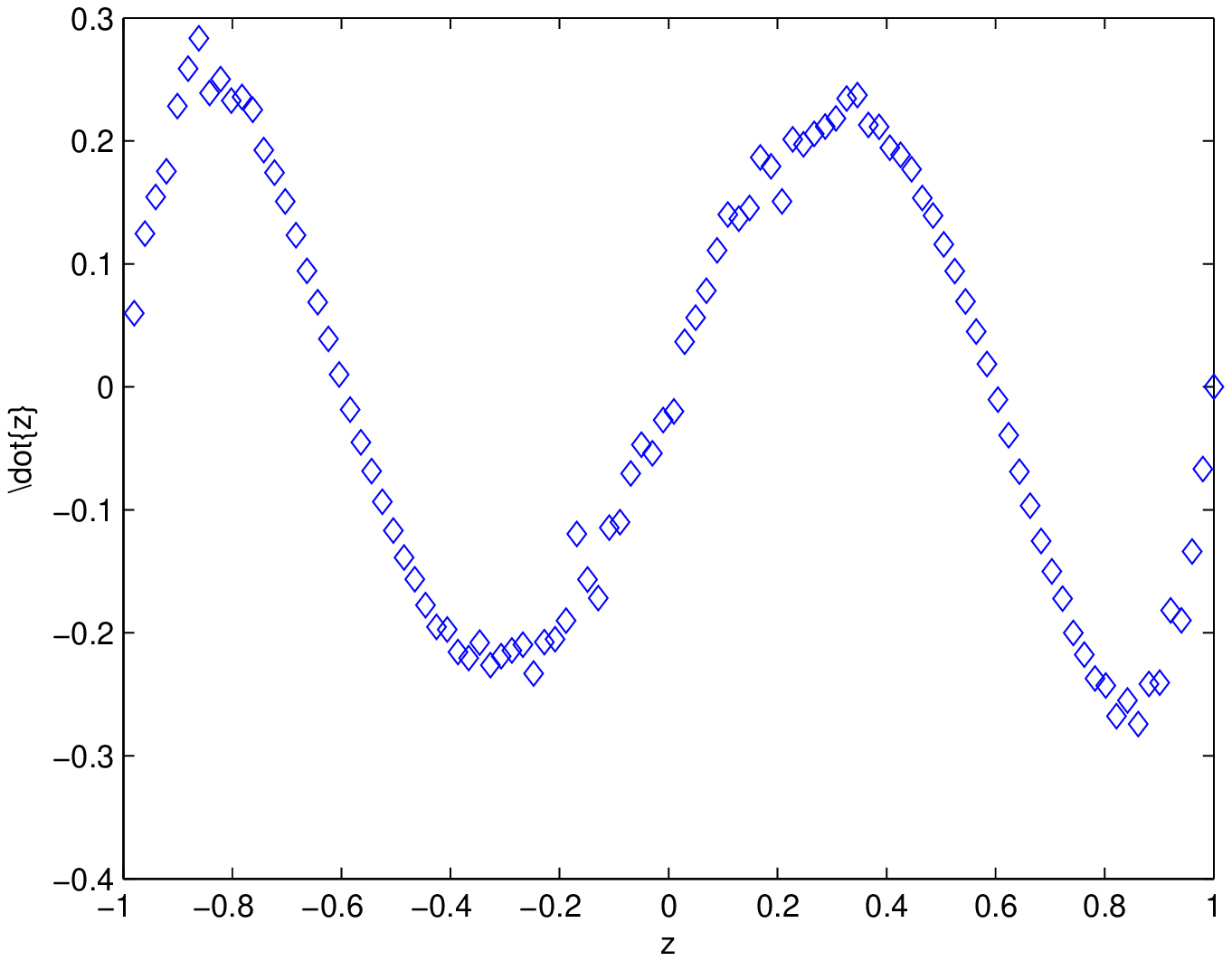}  
 \caption{Plot of $\dot{z}$ vs $z$ without noise according to rule~\ref{mMM}\label{fig:mMM}}
 \end{subfigure}
 \caption{}
\end{figure*}

\begin{figure}[h]
 \includegraphics[trim=0cm 0cm 0cm 0cm, clip=true, height=160 pt, width=0.5\textwidth]{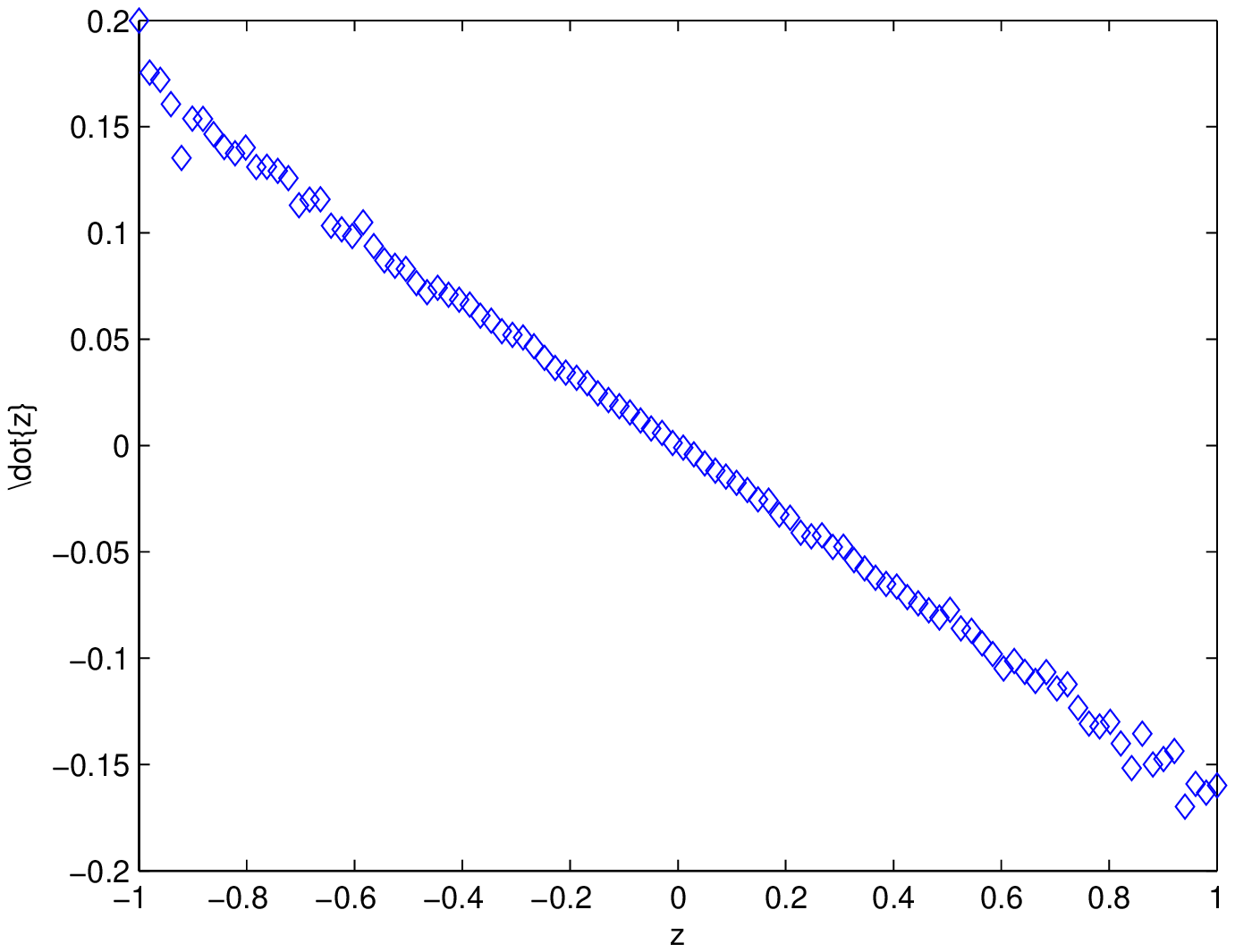}  
 \caption{Plot of $\dot{z}$ vs $z$ for purely noise driven system\label{pure_noise}}
\end{figure}

The figures~\ref{fig:MMM}-\ref{fig:mMM} show the different type of $\dot{z}$ vs $z$ curves arising from the different combinations of minority and majority rules. Now it can be clearly deduced that these curves are the weighted sum of the probability distributions of rule firing w.r.t. proper combination. When noise is introduced the contribution from noise gets added with the contributions from stochastic rule firing and the curve shifts. In every curve $z=\pm 1$ is a critical point, if the rule is majority dominated, those are stable and for minority dominated cases those are unstable. When noise is added the critical points are pushed inside. The plot of $\dot{z}$ vs $z$ for purely noise driven system is also given in the figure~\ref{pure_noise} and that constitutes the contribution from the noise component. It's almost straight linear with values equal to the noise level in both extrema. 

\section{Analytical Approach}

From the results of the experiments we obtained in the previous section, we can see a strikingly similar pattern between the probabilities of firing rules and the $\dot{z}$ vs $z$ trajectory of the Gillespie simulation. As in microscopic level we can model $\dot{z}$ as the average rate of change of $z$ with time or more precisely average change of $z$ in one time step and as $z$  changes by one step ($z\pm\dfrac{1}{N}$) only when a rule is fired or a noise driven transition takes place, we can conclude that $\dot{z}$  can be expressed as the weighted sum of the probabilities of firing different rules along with a term generated due to noise. So, our first concern is to find out the analytical expression for the aforementioned probabilities.

No, here during the implementation of the urn model we made $G$ draws with certain number of $X_1$(or $X_2$) decisions which may be termed as success from a finite population of N agents without replacement. So it satisfies the condition of hypergeometric distribution and getting $k$ numbers of $X_1$ decisions are termed as $k$ numbers of success. For a group size $G$, $k$ varies from $0$ to $G$. As for both the extreme cases there can't be any transition and therefore no realistic rule for $k=0$ and $G$, the total number of rules in a specific rule-set can be $G-1$. Again as each of the rule can be implemented as majority or minority, the total number of rule set can be $2^{\dfrac{G-1}{2}}$ where $G$ is an odd positive integer. The division by two takes place because rules must be symmetric to make the system realistic.

\begin{figure}[b]
 \includegraphics[trim=0cm 0cm 0cm 0cm, clip=true, height=160 pt, width=0.5\textwidth]{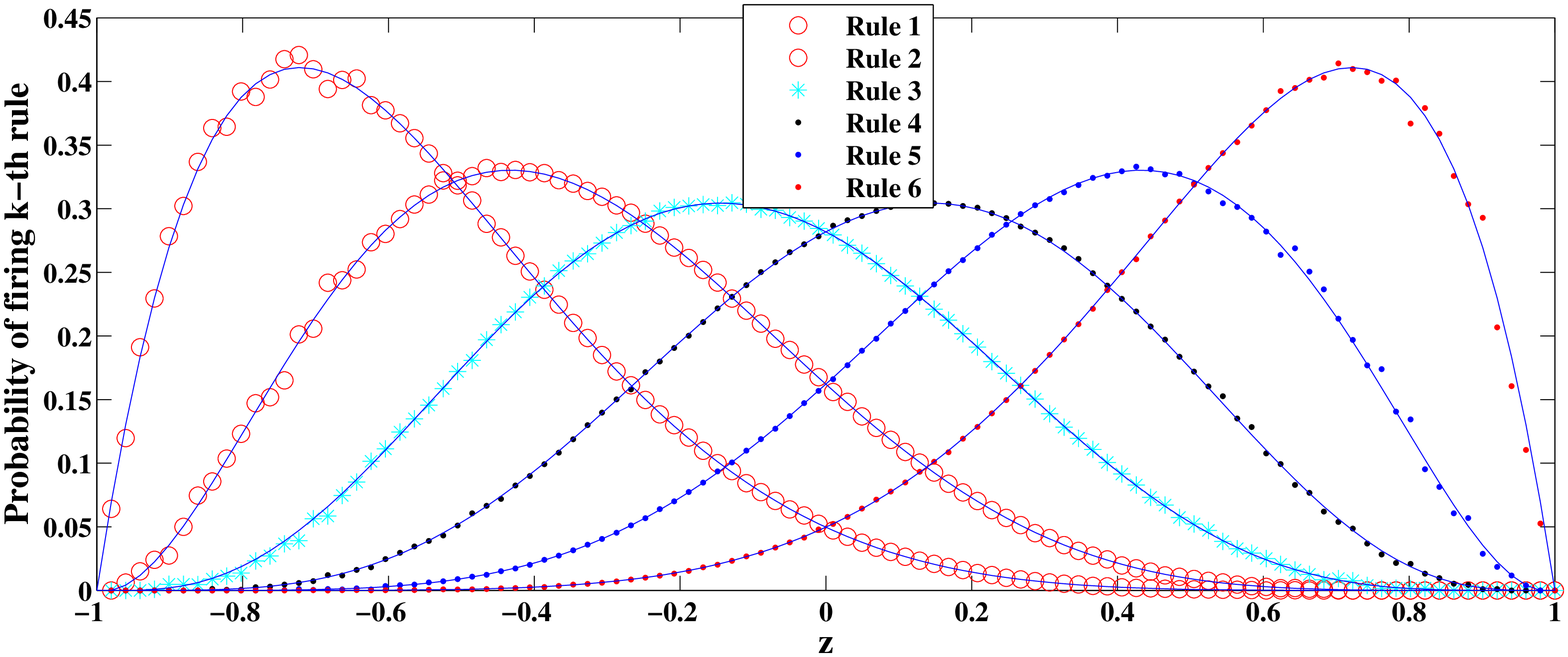}  
 \caption{Plot of theoretical and experimental values of the k-th rule's probability with different $k$ for $G=7$\label{hypergeometric}}
\end{figure}

So, with the convention that the probability of firing of the $k$-th rule is the probability of k successes within $G$ draws, and applying the probability mass function of hypergeometric distribution we get,
\begin{equation}
\label{eq:kth_probability}
P(X=k)=\dfrac{\binom{K}{k}\binom{N-K}{G-k}}{\binom{N}{G}},
\end{equation}
where $K$ is the number of success states in the whole population i.e. number of agents with decision $X_1$ among $N$ agents.
Now if we plot the equation~\ref{eq:kth_probability} vs $z$ for different $k$ and compare them with the experimental plot of the probabilities, we can see the concurrency between theoretical and experimental result as shown in the figure~\ref{hypergeometric}.

Now, as it is mentioned before, the expression for $\dot{z}$ can be written as the weighted sum of $P(X=k)$ $\forall k$ where weight is $\pm 1$ for majority and minority rule respectively. So, the expression of $\dot{z}$ for CDM without noise is given by

\begin{equation}
\label{eq:trajectory}
\dot{z}=\sum\limits_{\forall k}w_kP(X=k),
\end{equation}
where $P(X=k)$ is given by the equation~\ref{eq:kth_probability} and $K$ can be replaced by $N\dfrac{z+1}{2}$. $w_k$ is given by
\[w_k=\begin{cases}+1 & \mbox{if}\, k\mbox{-th rule is majority one}\\-1 & \mbox{if} \, k\mbox{-th rule is minority one.}\end{cases}\]
For a noise with uniform distribution (white noise) the contribution to $\dot{z}$ will be almost linear as shown in the figure~\ref{pure_noise} and as shown analytically by Bincalani \emph{et al} by the drift term $-z$. So for noisy system the equation~\ref{eq:trajectory} can be rewritten as 
\begin{equation}
\label{eq:trajectory_noise}
\dot{z}=-\epsilon z+\sum\limits_{\forall k}w_kP(X=k),
\end{equation}
where $\epsilon$ defines the noise level and all other symbols have their previous meanings.

\section{Conclusion}
In this project we tried to investigate the microscopic behaviour of a CDM process and relate it to a macroscopic description. We extend the approach of Bincalani \emph{et al} \cite{ref14} and Hamann \emph{et al} \cite{ref15}. In the former only the interaction between two agents and a purely noise-driven system is modelled where as in the latter one the model of using only majority rules is given. We experimentally determined the plot of the average change of the swarm fraction $\dot{z}$ and the probabilities of firing the majority and minority rules. We have shown that the drift trajectory is nothing but the weighted sum of these probability distributions which can be successfully modelled by hyper-geometric distribution and thus a purely analytic form of $\dot{z}$ can be given. 
This work can be extended in several ways. For example, we can search for a continuous analogue of hyper-geometric distribution to make the principle equation that governs a CDM process more compact. This search can lead to the discovery of certain orthogonal polynomials that can be used as the basis function to deterministically derive the coefficients like a frequency-domain transformation.

\vspace{10pt}
\section{Acknowledgements}
This project is supported by DAAD (Deutscher Akademischer Austauschdienst, German Academic Exchange Service) under WISE-2014 Fellowship programme. I want to thank my project supervisor Jun-Prof Dr. Heiko Hamann for his valuable advice and support that kept me running on this project. I also thank the members of Swarm Intelligence Group, University of Paderborn for their support and my home institute Jadavpur University, India to permit me continuing the project without slightest hindrance.


\begin{thebibliography}{99}
\bibitem{ref1}
Bonabeau, E., Dorigo, M., Theraulaz, G.: Swarm Intelligence: From Natural to
Artificial Systems. Oxford Univ. Press (1999)
\bibitem{ref2}
Schweitzer, F.: Brownian Agents and Active Particles. On the Emergence of Com-
plex Behavior in the Natural and Social Sciences. Springer-Verlag, Berlin, Germany
(2003)
\bibitem{ref3}
Alexander, J.C., Giesen, B., M ̈
unch, R., Smelser, N.J., eds.: The Micro-Macro
Link. University of California Press (1987)
\bibitem{ref4}
Schillo, M., Fischer, K., Klein, C.T.: The micro-macro link in DAI and sociology.
In Moss, S., Davidsson, P., eds.: Multi-Agent-Based Simulation: Second Interna-
tional Workshop, Boston, MA, USA (MABS 2000). Volume 1979 of LNCS., Berlin,
Germany, Springer-Verlag (2000) 303–317
\bibitem{ref5}
Hamann, H.: Space-Time Continuous Models of Swarm Robotics Systems: Sup-
porting Global-to-Local Programming. Springer-Verlag, Berlin, Germany (2010)
\bibitem{ref6}
Brambilla, M., Ferrante, E., Birattari, M., Dorigo, M.: Swarm robotics: a review
from the swarm engineering perspective. Swarm Intelligence 7(1) (2013) 1–41
\bibitem{ref7}
Prorok, A., Correll, N., Martinoli, A.: Multi-level spatial models for swarm-robotic
systems. The International Journal of Robotics Research 30(5) (2011) 574–589
\bibitem{ref8}
Berman, S., Kumar, V., Nagpal, R.: Design of control policies for spatially inho-
mogeneous robot swarms with application to commercial pollination. In LaValle,
S., Arai, H., Brock, O., Ding, H., Laugier, C., Okamura, A.M., Reveliotis, S.S.,
Sukhatme, G.S., Yagi, Y., eds.: IEEE International Conference on Robotics and
Automation (ICRA’11), Los Alamitos, CA, IEEE Press (2011) 378–385
\bibitem{ref9}
Franks, N.R., Mallon, E.B., Bray, H.E., Hamilton, M.J., Mischler, T.C.: Strategies
for choosing between alternatives with different attributes: exemplified by house-
hunting ants. Animal Behavior \textbf{65} (2003) 215–223
\bibitem{ref10}
Dussutour, A., Beekman, M., Nicolis, S.C., Meyer, B.: Noise improves collective
decision-making by ants in dynamic environments. Proceedings of the Royal Soci-
ety London B \textbf{276} (December 2009) 4353–4361
\bibitem{ref11}
Buhl, J., Sumpter, D. J. T., Couzin, I. D., Hale, J. J., Despland, E., Miller, E. R., Simpson, S. J.: From Disorder to Order in Marching Locusts. Science, 312(5778)(2006) 1402-1406
\bibitem{ref12}
Montes de Oca, M., Ferrante, E., Scheidler, A., Pinciroli, C., Birattari, M., Dorigo,
M.: Majority-rule opinion dynamics with differential latency: a mechanism for
self-organized collective decision-making. Swarm Intelligence \textbf{5} (2011) 305–327
\bibitem{ref13}
Valentini, G., Hamann, H., Dorigo, M.: Self-organized collective decision making:
The weighted voter model. In: 13th Int. Conf. on Autonomous Agents and Mutli-
agent Systems (AAMAS 2014).
\bibitem{ref14}
Biancalani, T., Dyson, L., McKane, A.J.: Noise-induced bistable states and their
mean switching time in foraging colonies. Phys. Rev. Lett. 112 (Jan 2014) 038101
\bibitem{ref15}
Hamann, H., Valentini, G., Khaluf, Y., Dorigo, M.: Derivation of a Micro-Macro Link
for Collective Decision-Making Systems: Uncover Network Features Based on Drift Measurements. 
\end{thebibliography}
\end{document}